\documentclass[conference]{IEEEtran}
\IEEEoverridecommandlockouts
\usepackage{booktabs}
\usepackage{multirow}
\usepackage{array}

\usepackage{cite}
\usepackage{amsmath,amssymb,amsfonts}
\usepackage{algorithmic}
\usepackage{graphicx}
\usepackage{textcomp}
\usepackage{xcolor}
\usepackage[table]{xcolor} 
\definecolor{green}{RGB}{200, 230, 200} 
\definecolor{red}{RGB}{255, 220, 220} 
\definecolor{blue}{RGB}{200,220,255} 
\usepackage{hyperref}

\def\BibTeX{{\rm B\kern-.05em{\sc i\kern-.025em b}\kern-.08em
    T\kern-.1667em\lower.7ex\hbox{E}\kern-.125emX}}
\begin{document}

\title{Dual-Interest Sequential Product Recommendation With Multi-Granular SSM\\
}

\author{
    \IEEEauthorblockN{Shuiying Liao}
    \IEEEauthorblockA{
        The Hong Kong University of Science and Technology\\
        Clear Water Bay, Hong Kong\\
        shuiyingl@ust.hk
    }
    \and
    \IEEEauthorblockN{P. Y. Mok$^{\ast}$}
    \IEEEauthorblockA{
        The Hong Kong University of Science and Technology\\
        Clear Water Bay, Hong Kong\\
        tracy.mok@ust.hk
    }
}

\maketitle

\begin{abstract}
Sequential recommendation aims to predict the next item a user will interact with based on their historical behavior. Advances in Transformers have significantly improved sequential recommendation but still limited by cost efficiency.  
Although State Space Models (SSMs) have recently enabled efficient long-range modeling, most existing methods models encode each item with a single static contextual role, overlooking the phenomenon of \textit{item polysemy}. In fact, the same item often plays different semantic roles depending on user context, and existing methods are limited in capture dynamic behavior across different temporal granularities.
In this work, we propose \textbf{DSRec}, a novel dual-interest cross SSM model that explicitly disentangles item roles across long-term and short-term semantic context. Sequential items are encoded into long-term interest embeddings that captures stable preferences via historical aggregation, and a short-term interest branch that emphasizes local session intent modulated by inter-click time intervals. These interest embeddings are processed through distinct SSM encoders: a full-sequence Mamba for long-term modeling, and a time-modulated SSM that dynamically adjusts state evolution based on temporal gaps.
To enable effective cross-granularity alignment, we adopt residual cross-fusion mechanism that exchanges contextual information between the two branches while preserving semantic independence. Experiments on public benchmarks demonstrate that DSRec outperforms other state-of-the-art methods. 
\end{abstract}


\section{Introduction}

Sequential recommendation has become a fundamental task in modern recommender systems, where the goal is to predict a user's next interaction based on their historical behavioral sequence~\cite{SASRec, bert4rec, liao2026consistency}. With the surge of online activities and the increasing volume of temporally ordered data, it has become crucial to model both \emph{long-term stable preferences} and \emph{short-term intent dynamics} to achieve personalized and timely recommendations~\cite{MBHT, liao2024hypergraph, liao2023recommendation}.
Transformer-based models, such as SASRec~\cite{SASRec} and BERT4Rec~\cite{bert4rec}, have shown promising results in modeling user sequences through attention mechanisms. However, their high computational complexity and limited scalability to long sequences hinder their practical deployment~\cite{Mamba, Jamba}. Recent developments in structured state space models (SSMs), including S4~\cite{gu2021efficiently}, S5~\cite{smith2022simplified}, and Mamba~\cite{Mamba}, provide efficient alternatives by enabling long-range dependency modeling with linear complexity. Consequently, SSMs are gaining increasing interest in recommendation tasks~\cite{Mamba4Rec, SIGMA}.

\begin{figure}[t]
  \centering
  \setlength{\abovecaptionskip}{0.2cm}
  \includegraphics[width=\linewidth]{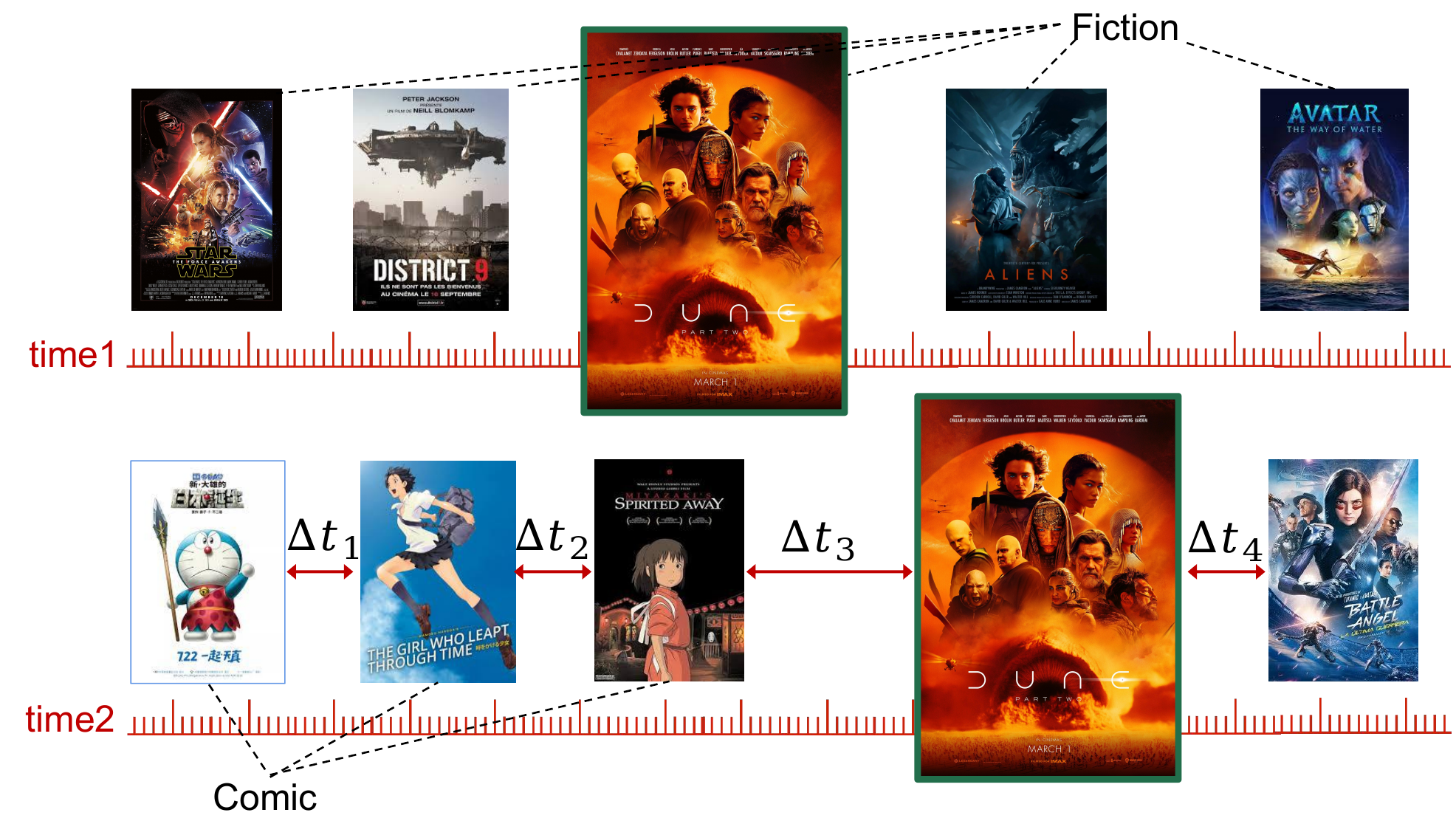}
  \caption{Motivation illustration. The same item (\textit{"Dune"}) can appear in different viewing sequences—either among consistent sci-fi films (long-term preference) or after comics (short-term interest). This demonstrates the polysemous nature of item semantics, requiring recommendation to distinguish temporal contexts for accurate personalization. }
  \label{fig: example}
\end{figure}

Despite their advantages, most existing SSM-based recommender models treat user behavior as a homogeneous sequence and apply a single representation for each item across all contexts. However, user-item interactions are inherently multi-intent: a user may engage with the same item due to different motivations. In fact, \textit{an item can carry polysemous semantics roles.} Take Fig.~\ref{fig: example} as an example, the same item (e.g., \textit{Dune}) may occur in different contexts, appearing alongside other science fiction films as part of a user's long-term preference, or following a sequence of comics driven by recent short-term interest. This illustrates the polysemous nature of item semantics and highlights the necessity of modeling different interest intent to enable accurate personalization.

Although prior works have explored multi-interest modeling~\cite{MBHT, cl4srec, duorec}, they often focus on user-side intent separation or perform sequence slicing at the cost of disrupting temporal coherence~\cite{liao2024hypergraph, liao2026hamiltonian, liao2026pcgnet}. These limitations motivate us to design a model that captures item-level semantic disambiguation across multiple temporal granularities.

To address these limitations, we propose \textbf{D}ual \textbf{S}SM \textbf{Rec}ommendation (\textbf{DSRec}), a novel dual-interest sequential recommendation model based on multi-granular cross residual SSM encoders. Specifically, we first adopt a dual-interest modeling that explicitly distinguish the short-term and long-term semantic roles of each item. 
Each item in the sequence is embedded into two semantic roles: a \textit{long-term interest} embedding that encodes persistent user preferences through historical aggregation, and a \textit{short-term interest} modeling that emphasizes local session intent modulated by inter-click time intervals. The long-term and short-term interest are independently encoded using structurally distinct SSM-based architectures: one using a standard full-sequence Mamba \cite{Mamba4Rec}, and the other using a time-modulated SSM that dynamically adjusts state updates based on time gaps.

Importantly, as using two identical Mamba encoders overlooks the distinct temporal patterns present in short-term and long-term behaviors. Long-term preferences benefit from full-sequence modeling and high-order memory, while short-term interest depends heavily on session-local dynamics and time sensitivity. Hence, we design asymmetric SSM encoders, to optimize each path for its respective granularity.
Rather than merging the two branches via simple concatenation or early fusion, we adopt a residual cross-fusion mechanism that allows each branch to inject contextual signals from the other without disrupting their specialized semantics. This cross-residual design promotes joint modeling of multi-granular dependencies. 

Extensive experiments on benchmark datasets show that DSRec outperforms state-of-the-art methods in terms of both accuracy and efficiency. Ablation studies further verify the effectiveness of key modules. We believe DSRec opens a new path for item role-aware modeling in sequential recommendation, which has not been explored in existing sequential recommendation systems.

\vspace{2mm}




Our main contributions are as follows:
\begin{itemize}
    \item We propose item polysemy of long-term interest and short-term interest semantical roles in sequential recommendation. We adopt polysemous modeling that encodes each item into different interest patterns, enables the model to handle item semantic ambiguity across diverse user histories and session contexts.
    

    \item We design an \textit{asymmetric dual-SSM architecture} composed of a global Mamba encoder for modeling long-term user interest and a time-modulated SSM for capturing short-term dynamics. A residual cross-fusion mechanism is adopted to enable bidirectional information exchange between the two branches while preserving their semantic independence.

    \item We extensively evaluate our model on multiple public benchmarks, demonstrating that our approach surpasses existing methods in terms of overall accuracy and efficiency.
\end{itemize}

\section{Related Work}
\subsection{Sequential Recommendation}
Sequential recommendation aims to predict the next item a user will interact with based on their historical behavior sequence \cite{liao2024hypergraph, ding2021leveraging, ding2021modeling}. Early research in sequential recommendation systems heavily relied on RNNs \cite{hidasi2015session, li2017neural}, which were adept at handling sequential data due to their ability to maintain a hidden state across different time steps. However, RNNs faced limitations in parallel computation and struggled with capturing long-range dependencies within sequences. These challenges led to the exploration of Transformer-based models \cite{SASRec, bert4rec, duorec, maerec, dcrec, cl4srec, iclrec, liao2024hypergraph}, which introduced self-attention mechanisms to overcome the sequential processing constraints. Several works have attempted to enhance performance via self-supervised pretraining~\cite{zhou2020s3} or dual representations~\cite{duorec, liao2024reproducibility}.
However, despite the improved performance, the quadratic computational complexity of Transformers with respect to sequence length posed a challenge, especially in real-time recommendation scenarios where latency is critical \cite{fan2024tim4rec}. The need for more computationally efficient models in sequential recommendation systems has driven the adoption of SSMs \cite{Mamba, huang2025localmamba, SIGMA, zhang2024matrrec, zhang2025local, liu2024bidirectional}. 
These models offer linear computational loads, making them more suitable for handling long sequences and real-time recommendations \cite{gu2020hippo, peng2023rwkv}.

\textit{Our work complements these trends by exploring a new modeling dimension: semantic interest roles disentanglement, which explicitly separates short- and long-term item semantics and models them via dual SSMs for efficient and more accurate sequence learning.}


\subsection{State Space Models.}
State Space Models (SSMs) \cite{hamilton1994state, aoki2013state} have recently gained popularity in sequential modeling due to their ability to capture long-range dependencies with linear computational complexity \cite{Mamba, huang2025localmamba, qu2024ssd4rec}. In contrast to the quadratic cost of Transformers, SSMs leverage recurrence-based mechanisms to process sequences efficiently, making them especially attractive for long-context scenarios \cite{Mamba, Jamba}.
Mamba \cite{Mamba} first introduces a selective state space model with hardware-aware design, marked a breakthrough in combining high efficiency with strong sequence modeling. Following this, a variety of models extended Mamba to other domains. For example, Lieber et al.~\cite{Jamba} proposed Jamba, a hybrid Transformer-Mamba large language model that supports efficient long-context processing through a mixture-of-experts architecture. In the vision domain, Wu and Wan~\cite{wu2025medical} and Liu et al.~\cite{liu2024vmamba} applied SSMs to image segmentation, demonstrating their superiority in capturing global spatial dependencies over Transformer-based approaches.
In sequential recommendation domain, Mamba4Rec \cite{Mamba4Rec} introduces the first efficient selective state-space architecture, provides a robust solution for capturing complex temporal dynamics in user sequential behavior.
To enhance SSMs further, SIGMA~\cite{SASRec} extended Mamba with selective gating and dense extraction layers to better adapt to short sequences and unidirectionaly sequence modeling, and shown promising results.

\textit{However, existing Mamba-based methods in recommendation mostly treat items as a static semantic role and apply monolithic architectures for modeling sequence dynamics.} 


\begin{figure*}[t!]
  \centering
  \setlength{\abovecaptionskip}{0.2cm}
  \includegraphics[width=0.9\linewidth]{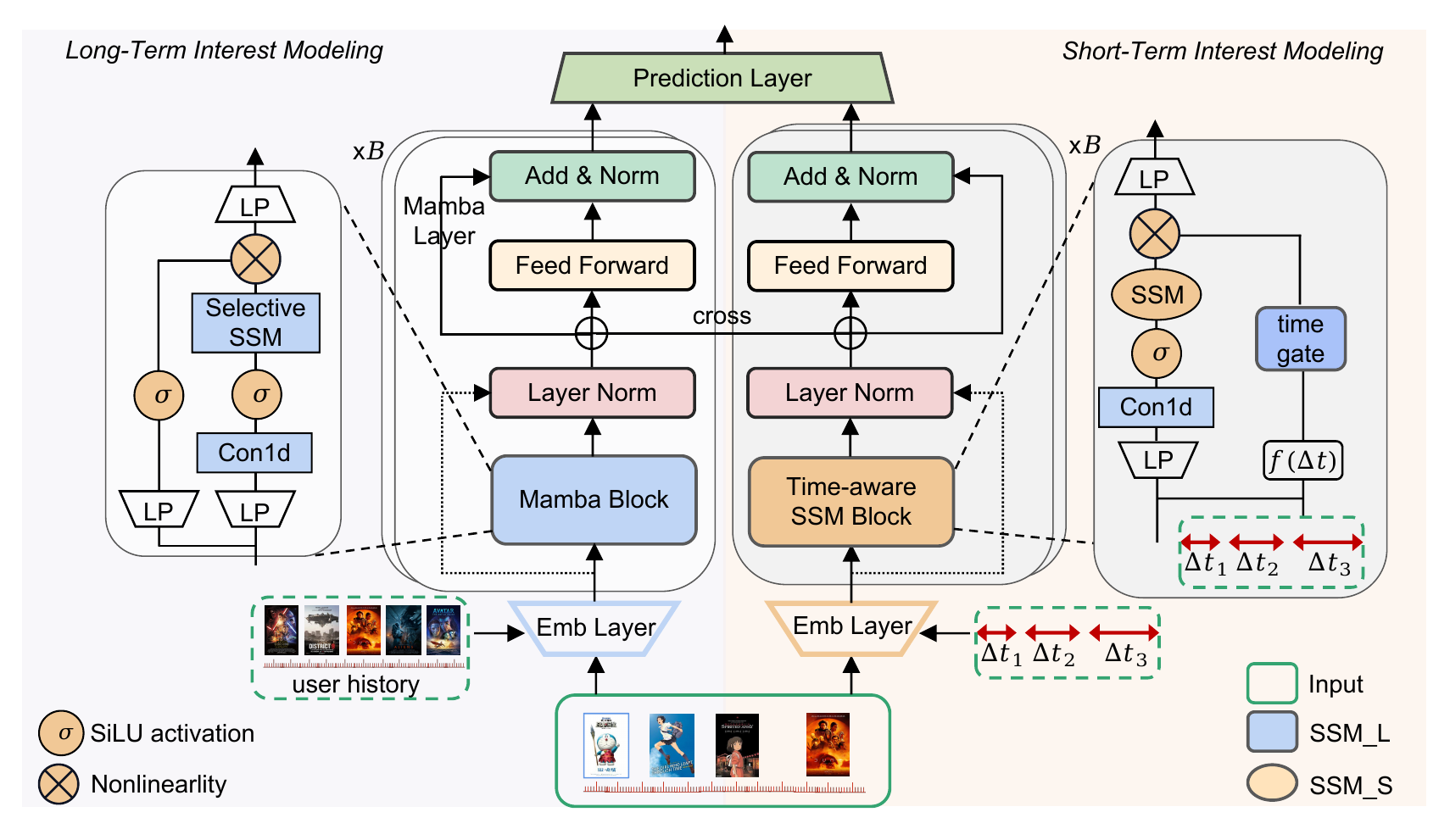}
  \caption{Structure overview. DCRec is a dual-SSM architecture that explicitly disentangles short-term and long-term user preferences via embedding and state-space separate modeling. The two streams are connected through residual cross-links.}
  \label{fig: main}
\end{figure*}

\section{Methodology}

\subsection{Notations Problem Statement}
Let $\mathcal{U}$ denotes a set of users and $\mathcal{V}$ denotes a set of items, where $|\mathcal{U}|$ and $|\mathcal{V}|$ denote the number of users/items, $u \in \mathcal{U}$ and $v \in \mathcal{V}$. The sequential recommendation aims to predict the next item that a user $u$ is likely to interact with based on their historical sequence $S_u$ of interactions. Therefore, we represent a historical sequence of interactions for the user $u$ with $S_u = \left[v_1,v_2, \ldots, v_T \right]$, where $v_t$ represents the item interacted with the user $u$ at time step $t$, and here we omit the subscript $u$ for convenience. $v_{T+1}$ represents the next item that user $u$ is expected to interact with, which is to be predicted.

Based on the above notations, given a set of historical interaction sequences $S_u$ for users in the training dataset $\mathcal{D}=\left\{\left(S_u, v_{T+1}\right)\right\}$, the task of sequential recommendation is to learn a model 
$\mathcal{F}: S_u \rightarrow v_{T+1}$ that can accurately predict the next item $v_{T+1}$ for each user $u$.
\vspace{2mm}

\subsection{Framework Overview}
In \textbf{DSRec}, we model long-term interest and short-term interest jointly. We adopt dual-SSM sequential recommendation model based on residual-coupled state space networks. Each item in the input sequence is represented by two contextual embeddings: a \textit{long-term embedding} that captures persistent preferences from historical interactions, and a \textit{short-term embedding} that emphasizes local intent modulated by inter-click time gaps.
The two contextual embeddings are processed independently through dedicated SSM encoders: the long-term embedding is modeled with standard Mamba \cite{Mamba}, while the short-term embedding is processed using a time-modulated state-space model. To enable effective coordination, a residual cross-fusion mechanism is introduced between branches. The fused representations are then projected for next-item prediction. The main structure are illustrated in Fig. \ref{fig: main}. 

\vspace{2mm}

\subsection{State Space Model}
State Space Models (SSMs) are a class of dynamical systems widely used for modeling temporal dependencies in continuous or discrete-time signals. Recent variant Mamba has shown strong performance in sequence modeling tasks 
due to their efficient handling of long-range dependencies.
Formally, an SSM defines the evolution of a hidden state $h(t)$ over time based on an input signal $x(t)$, and generates an output signal $y(t)$ through the following continuous-time equations:
\begin{align}
\frac{d}{dt} h(t) &= \mathbf{A} h(t) + \mathbf{B} x(t) \label{eq:ssm_continuous1} \\
y(t) &= \mathbf{C} h(t) 
\label{eq:ssm_continuous2}
\end{align}
where $\mathbf{A} \in \mathbb{R}^{P \times P}$, $\mathbf{B} \in \mathbb{R}^{P \times D}$, and $\mathbf{C} \in \mathbb{R}^{D \times P}$ are learnable parameter matrices, and $P$ denotes the dimension of the hidden state space.
To apply SSMs in discrete-time sequential recommendation, we discretize the above system using the Zero-Order Hold (ZOH) method with a constant step size $\Delta t$, resulting in:
\begin{align}
\mathbf{h}_t &= \bar{\mathbf{A}} \mathbf{h}_{t-1} + \bar{\mathbf{B}} \mathbf{x}_t \label{eq:ssm_discrete1} \\
\mathbf{y}_t &= \mathbf{C} \mathbf{h}_t \label{eq:ssm_discrete2}
\end{align}
Here, $\bar{\mathbf{A}}$ and $\bar{\mathbf{B}}$ are the discretized equivalents of $\mathbf{A}$ and $\mathbf{B}$, typically computed via matrix exponentials or approximation methods. The model sequentially updates the internal state $\mathbf{h}_t$ given an input embedding $\mathbf{x}_t$, and produces the output $\mathbf{y}_t$ at each time step. This formulation allows the model to capture structured temporal dynamics in user-item interaction sequences. For simplicity, we shorten the whole process to SSM in the following two branches.

\subsection{Long-Term Interest Modeling}

\subsubsection{Embedding Layer} 
We first establish a learnable embedding table $\mathbb{E}=\left\{E_1, E_2, \cdots, E_{|\mathcal{V}|}\right\} \in \mathbb{R}^{\mathcal{|V|}\times\mathrm{D}}$. Each item $v_j \in \mathcal{V}$ is initially represented as a one-hot vector and then projected into a dense vector space $\mathbb{R}^{|\mathcal{V}| \times D}$ via an embedding matrix $\mathbb{E}$. Given an interaction sequence $\mathcal{S} = \{v_1, v_2, \ldots, v_T\}$, the corresponding embedded input is denoted as $\boldsymbol{X} \in \mathbb{R}^{T \times D}$, obtained by retrieving embeddings from $\mathbb{E}$.

\vspace{2mm}

\textit{User History Aggregation for Long-Term Interest.}  
To capture consistent user preferences beyond the current session, we enhance each item's long-term interest embedding with a user-level historical aggregation $\boldsymbol{X}_l =MLP(\boldsymbol{X}||E_u)$. The intuition is that long-term interests are not determined by the current session alone, but by its relative semantics within a user's broader interaction history $E_u$. We compute this aggregation by applying a mean pooling over all previous items in the user's sequence, excluding the current item, and use it as a contextual shift added to the item's base embedding:
\begin{equation}
    \mathbf{x}_l^i = \mathbf{x}_i + \frac{1}{i-1} \sum_{j=1}^{i-1} \mathbf{x}_j,
\end{equation}
where $\mathbf{e}_i$ is the embedding of the $i$-th item and the second term captures aggregated historical context. This design allows the long-term encoder to better model stable preferences and suppress short-term fluctuations.

To enhance the expressiveness and generalization of the item embeddings, we then apply dropout regularization~\cite{dropout} and layer normalization~\cite{layernorm} on the input embbeding $\boldsymbol{X}_l$ for long-term interest branch.  
These operations mitigate overfitting and stabilize training dynamics, ensuring a more robust embedding space for downstream modeling.

\vspace{2mm}

\subsubsection{Mamba Block}
To capture long-range user behavior efficiently, we keep and leverage structured SSMs as Mamba4Rec \cite{Mamba4Rec} dose. The core component is the \textbf{Mamba Block}, which transforms an input sequence $\boldsymbol{X}_l \in \mathbb{R}^{B \times L \times D}$ (with batch size $B$, sequence length $L$, and hidden dimension $D$) into output features using a selective SSM kernel and residual connections.

Each Mamba Block transforms an input sequence tensor $\boldsymbol{H}_i \in \mathbb{R}^{B \times L \times D}$, where $B$ is the batch size, $L$ is the sequence length, and $D$ is the hidden dimension. The transformation follows a structured state space modeling pipeline with selective parameterization.
First, the input undergoes a linear projection and 1D convolution to capture local patterns, followed by a gated activation:
\begin{equation}
\boldsymbol{H}_x = \text{SiLU}(\text{Conv1D}(\boldsymbol{W}_1 \boldsymbol{X}_l)),
\label{conv}
\end{equation}
where $\boldsymbol{W}_1 \in \mathbb{R}^{D \times D}$ is a learnable projection matrix, and $\text{SiLU}(\cdot)$ is the Sigmoid Linear Unit activation defined as $\text{SiLU}(x) = x \cdot \sigma(x)$.
Next, the convolved features $\boldsymbol{H}_x$ are passed through a parameterized structured state space model (SSM), which applies input-dependent filters using learned state matrices $\boldsymbol{A}$, $\boldsymbol{B}$, and $\boldsymbol{C}$:
\begin{equation}
\boldsymbol{H}_y = \text{SSM}(\boldsymbol{H}_x; \boldsymbol{A}, \boldsymbol{B}, \boldsymbol{C}),
\label{ssm}
\end{equation}
where $\text{SSM}(\cdot)$ denotes the discretized recurrence function modeling temporal dynamics.

To ensure gradient flow and preserve original semantics, a residual connection is introduced with a second projection:
\begin{equation}
\boldsymbol{H}_l = \boldsymbol{W}_2 (\text{SiLU}(\boldsymbol{H}_y)) + \boldsymbol{X}_l.
\label{lp}
\end{equation}
where $\boldsymbol{W}_2$ is another trainable linear layer. This operation fuses the SSM-refined representation with the original input for stable layer-wise integration.
The Mamba Block jointly leverages local convolutional processing and global state space modeling, while the use of input-dependent parameters and residual connections facilitates efficient long-sequence encoding in recommendation scenarios.

\subsection{Short-Term Interest Modeling}
\subsubsection{Embedding Layer}
To account for the multi-role nature of items in user behavior sequences, we first assign each item two learned embeddings: $\boldsymbol{X}_l$ is assigned as \textit{Long-term interest embedding}, combined with user history shift as we mentioned. While in short-term interest modeling, we first use $\boldsymbol{X}_s$ represents \textit{Short-term interest embedding} after a  projection from initial sequence embedding $\boldsymbol{X}$, to encode the transient intent and session-specific semantics that may vary across occurrences. Both representations are learned independently.

As user short-term intent in recommendation sequences is often influenced by the timing of recent interactions. Intuitively, two items clicked within a short time span are more likely to be semantically coherent than those separated by long intervals. To capture such dynamics, we integrate \textbf{time interval encoding} into the short-term interest modeling branch.

\vspace{2mm}

\textit{Time Interval Encoding:}
To incorporate temporal dynamics into sequential modeling, we process the timestamp sequence $\mathcal{T} = \{t_1, t_2, \ldots, t_T\}$ by computing its temporal difference sequence $\mathcal{D} = \{d_1, d_2, \ldots, d_T\}$ \cite{fan2024tim4rec}, where each $d_i$ represents the elapsed time between inter-click intervals. Specifically, the differences are computed as follows:
\begin{equation}
\mathcal{D}_{i} = 
\begin{cases}
0, & i=1 \\
\mathcal{T}_{i} - \mathcal{T}_{i-1}, & i = 2, \ldots, T
\end{cases}.
\end{equation}
The resulting sequence $\mathcal{D} \in \mathbb{R}^{T}$ captures fine-grained temporal intervals between actions. Then we assign each time interval a learnable embedding. To reduce variance and suppress temporal noise, we further apply dropout and layer normalization to $\mathcal{D}$ prior to feeding it into downstream temporal modules. This enables the model to better distinguish between long pauses and rapid interactions, improving its sensitivity to user intent shifts over time.
The raw intervals are then normalized via logarithmic scaling and discretized into $N$ buckets using quantile binning. Each bucket is mapped to a learnable embedding $\mathbf{x}_i^{\text{t}}$, capturing temporal sensitivity. This time encoding is concatenated with the item embedding to form the short-term interest representation:

\begin{equation}
    \mathbf{x}_s^i = \mathrm{MLP}_S\left(\mathrm{x}_i \, \Vert \, \mathbf{x}_i^{\text{t}} \right),
\end{equation}


\vspace{2mm}

\subsubsection{Time-aware SSM}

To further enhance temporal sensitivity, we introduce a gating mechanism that modulates the state update based on time interval after similar process as Eq. \ref{conv}, \ref{ssm}, and \ref{lp}:
\begin{equation}
\mathbf{g}_i = \sigma(\boldsymbol{W}_t \cdot \mathbf{x}_i^{\text{t}}),
\end{equation}
\begin{equation}
\mathbf{h}^i_s = \mathbf{g}_i \odot \text{SSM}(\mathbf{x}_s^i) + (1 - \mathbf{g}_i) \odot \mathbf{h}^{i-1}_s,
\end{equation}
where $\sigma(\cdot)$ is a sigmoid activation. The gate vector $g_i$ controls how much the current input should influence the hidden state. When the time gap is small, $g_i$ tends to be higher, allowing new information to be integrated; for large gaps, $g_i$ is smaller, retaining past state memory, allowing dynamic temporal sensitivity.
This time-aware control mechanism is critical for modeling transient interests that shift rapidly within user sessions.

\subsection{Residual Cross-SSM Fusion}
We construct two independent SSM encoders: \textit{Mamba Block} processes the $\boldsymbol{X}_l$ sequence to extract long-term user preference; \textit{Time-aware SSM} processes the $\boldsymbol{X}_s$ sequence to model session-level short-term dynamics. Each encoder is augmented with a residual cross-connection, meaning each branch receives the detached output of the other as auxiliary input. This enables information sharing while preventing entanglement of gradients.


\subsection{Feed-Forward Network}
The standard feed-forward network is adopted to enhance the modeling of user interactions in both long- and short-term interest latent space, which is defined as follows:

\begin{equation}
    \text{FFN}(H_l) = \text{GELU}\left(H_l W^{(1)} + b^{(1)}\right) W^{(2)} + b^{(2)},
\label{eq: ffn}
\end{equation}
\begin{equation}
    \text{FFN}(H_s) = \text{GELU}\left(H_s W^{(3)} + b^{(3)}\right) W^{(4)} + b^{(4)},
\end{equation}
where $W^{(1)},W^{(3)}  \in \mathbb{R}^{D \times 4D}$, $W^{(2)}, W^{(4)} \in \mathbb{R}^{4D \times D}$, $ b^{(1)},b^{(3)} \in \mathbb{R}^{4D} $, and $ b^{(2)},b^{(4)} \in \mathbb{R}^{D} $ are the parameters of two dense layers, and we utilize the GELU activation function \cite{gelu}.

The FFN is designed to capture complex patterns and interactions within sequential data by applying two non-linear transformations using dense layers with the activation function. To enhance model robustness and prevent overfitting, we incorporate dropout and layer normalization after each Mamba block/Time-aware SSM block and feed-forward network, as depicted in Eq. \ref{eq: ffn}. This approach aids in regularizing the model and accelerating training convergence.

\subsection{Stacking SSM Layers}
We investigate the use of stacked SSM layers ($B$) to enhance sequential recommendation. Despite deeper networks not always improving performance, residual connections are vital for propagating features across layers. We explore configurations for stacked SSM layers to balance effectiveness and efficiency. Further experiments are conducted to evaluate the trade-offs between effectiveness and efficiency of stacked layers.

\subsection{Prediction Layer and Optimization}
The final output embedding is formed by concatenating the last positions of long-term interest modeling output $h_t$ and long-term interest modeling output $s_t$:
\begin{equation}
    \mathbf{o}_T = \mathrm{MLP}([\mathbf{h}_{l}^T \, \Vert \, \mathbf{h}_{s}^T]).
\end{equation}
Prediction is computed via dot product over the item embedding matrix:
\begin{equation}
    \hat{y} = \mathrm{Softmax}(\mathbf{o}_T \cdot \mathrm{E}^\top) \in \mathbb{R}^{|\mathcal{V}|},
\end{equation}
where $\hat{y}$ is the probability distribution over the next item in the item set $\mathcal{V}$.
To optimize the DSRec, we adopt the cross entropy loss function as below:
\begin{equation}
\mathcal{L}=-\sum_{u \in \mathcal{U}} \sum_{i=1}^N y_{T+1} \log \left(\hat{y}_{T+1}\right)
\end{equation}
where $y_i$ represents the ground truth of item $v_{T+1}$.

\section{Experiments}
\begin{table*}[t]
\centering
\caption{Dataset statistics.}
\label{tab:dataset_stats}
\begin{tabular}{lcccccc}
\toprule
\textbf{Dataset} & \textbf{\# Users} & \textbf{\# Items} & \textbf{\# Interactions} & \textbf{Avg. actions of users} & \textbf{Avg. actions of items}& \textbf{Sparsity} \\
\midrule
MovieLens-1M & 6,041 & 3,417 & 999,611 & 165.5 & 292.6 & 95.157\%\\
Amazon-Beauty & 22,364 & 12,102 & 198,502 & 8.9 &16.4 & 99.927\%\\
Amazon-Video-Games & 24,304 & 10,673 & 231,780 & 9.5 &21.7 & 99.911\%\\
\bottomrule
\end{tabular}
\end{table*}

\vspace{2mm}

In this section, we first give brief introduction of the experimental setting, and to evaluate the effectiveness of our proposed DSRec, we conducted extensive experiments on two benchmark datasets, giving answers to the following research questions (RQs):

\vspace{2mm}

\begin{itemize}
    \item \textbf{RQ1:} Can the proposed DSRec outperform previous sequential recommendation methods?
    \item \textbf{RQ2:} Can the key modules contribute to the overall prediction accuracy?
    \item \textbf{RQ3:} How do specific settings in the method affect overall performance?
    \item \textbf{RQ4:} How does our model benefit sequential recommendations?
\end{itemize}

\subsection{Experimental Settings}

\subsubsection{\textbf{Data}}
We conduct our experiments on three public datasets, and all the datasets are collected from various real-world application platforms, and these datasets show considerable variation in both sequence length and sparsity.

\vspace{2mm}

\textbf{Beauty} and \textbf{Video-Games} \cite{mcauley2015image} are obtained from Amazon e-commerce platform, and the two dataset typically contains information related to beauty and movie products.

\textbf{MovieLens-1M} \cite{harper2015movielens} is a benchmark movie recommendation dataset which contains bout one million movie ratings from users.

\vspace{2mm}

The important statistics for the three datasets can be found in Table~\ref{tab:dataset_stats}.

\vspace{2mm}



\subsubsection{\textbf{Baselines}}
A variety of representative methods in sequential recommendation have been chosen for comparison against our approach. They are organized into the following distinct categories:

\vspace{2mm}

\paragraph{CNN based method} 
\begin{itemize}
\item \textbf{Caser} \cite{Caser}: It utilizes a convolutional layer to extract features from user interaction sequences, capturing both short-term and long-term patterns. 
\end{itemize}

\vspace{2mm}

\paragraph{RNN based method} 
\begin{itemize}
\item \textbf{GRU4Rec} \cite{hidasi2015session}: It refines conventional RNNs, incorporating a ranking loss function to achieve superior performance tailored to sequential recommendation domain.
\end{itemize}

\vspace{2mm}

\paragraph{Attention based methods} 
\begin{itemize}
\item \textbf{NARM} \cite{li2017neural}: This model employs an attention-based blended encoder to distill user intent from a sequence of actions, forming a unified session representation. 
\item \textbf{SASRec} \cite{SASRec}: It is a self-attention-based sequential model that harmonizes long-term semantic comprehension with a focus on a select set of pertinent actions. It dynamically selects pertinent past items at each step to forecast the subsequent item.
\item \textbf{BERT4Rec} \cite{bert4rec}: It further applies the Transformer model to capture long-term dependencies in sequences through a bidirectional transformer architecture, thus improving the accuracy and effectiveness. 
\end{itemize}

\vspace{2mm}

\paragraph{SSM based methods}
\begin{itemize}
\item \textbf{Mamba4Rec} \cite{Mamba4Rec}:It is the first to apply selective State Space Models (SSMs) in sequential recommendations, particularly the hardware-optimized Mamba block. \textit{This} method has attracted wide attention and applications.
\item \textbf{SIGMA} \cite{SIGMA}: This model is a hybrid model that integrates Mamba and GRU. It incorporates a Partially Flipped Mamba component combined with a Dense Selective Gate.
\end{itemize}

\vspace{2mm}

\subsubsection{\textbf{Implementation}}
Our evaluation is implemented based on PyTorch \footnote{\url{http://pytorch.org}} and RecBole \footnote{\url{https://github.com/RUCAIBox/RecBole}}. All models utilize a model dimension of 64, a standard selection for sequential recommendation models that focus solely on ID-based interactions. The optimizer is Adam \cite{Adam} with a learning rate of 0.001, and the training batch size is set to 2048. 
For Mamba-based models, we use structured state space model configurations with a state dimension was consistently set to 32, a local convolution width set to 4, the kernel size for 1D causal convolution was 4, and a block expansion factor was set to 2. We apply a dropout rate of 0.2 after each layer to prevent overfitting. All experiments are conducted on NVIDIA GeForce RTX 4090 GPUs. We employ a training batch size of 2048 and a validation batch size of 4096 across all experiments. The maximum sequence length is dynamically determined based on each dataset's characteristics: 200 for MovieLens-1M and 50 for both Amazon-Beauty and Amazon-Video-Games. All models are carefully tuned to achieve their best potential performance under unified experimental setup.

\vspace{2mm}

\subsubsection{\textbf{Evaluation}}
Followed by previous studies \cite{bert4rec, SASRec, Linrec, Mamba4Rec}, we employ the leave-one-out strategy to assess the effectiveness of each method. Under this strategy, for each user, the last interacted item in temporal order is considered as the test data and the previous one as validation data.
To ensure robust evaluation metrics, we utilize Hit Ratio (\textit{HR@}), Normalized Discounted Cumulative Gain (\textit{NDCG@}), and Mean Reciprocal Rank (\textit{MRR@}) to evaluate the performance of each method, as these metrics are widely accepted in the field of recommendation. In this study, we report the evaluation metrics at rank 10.

\subsection{Overall Performance}

\begin{table*}[t!]
\centering
\setlength{\tabcolsep}{2mm}
\caption{Overall Performance Comparison of Different Methods. \colorbox{red}{Red} and \textbf{Bold} data indicate the best results, and \colorbox{blue}{blue} \underline{\textit{underlined}} ones are the second best results. "$\star$" indicates the improvements are statistically significant}
\label{tab:performance}
\begin{tabular}{lccccccccc}
\toprule
\multirow{2}{*}{\textbf{Method}} & \multicolumn{3}{c|}{\textbf{MovieLens-1M}} & \multicolumn{3}{c|}{\textbf{Amazon-Beauty}} & \multicolumn{3}{c}{\textbf{Amazon-Video-Games}} \\
\cmidrule(lr){2-4} \cmidrule(lr){5-7} \cmidrule(lr){8-10}
 & \textbf{HR@10} & \textbf{NDCG@10} & \textbf{MRR@10} & \textbf{HR@10} & \textbf{NDCG@10} & \textbf{MRR@10} & \textbf{HR@10} & \textbf{NDCG@10} & \textbf{MRR@10} \\
\midrule
\textbf{Caser} \cite{Caser}~\textsuperscript{(WSDM'18)}& 0.2892 &0.1714 & 0.1354 &0.0531 & 0.0294&0.0222 &0.0891&0.0460&0.0330\\
\midrule
\textbf{GRU4Rec} \cite{hidasi2015session}~\textsuperscript{ICLR'16)} & 0.2934 & 0.1642 & 0.1249 & 0.0606 & 0.0332 & 0.0249 & 0.1030 & 0.0536 & 0.0380 \\
\midrule
\textbf{NARM} \cite{li2017neural}~\textsuperscript{(CIKM'17)}& 0.2735 & 0.1506 & 0.1132 & 0.0627 & 0.0347 & 0.0262 & 0.1032 & 0.0530 & 0.0379 \\
\textbf{SASRec} \cite{SASRec}~\textsuperscript{(ICDM'18)}& 0.2977 & 0.1687 & 0.1294 & \cellcolor{blue}\underline{\textit{0.0847}} & 0.0425 & 0.0296 & 0.1168 & 0.0571 & 0.0390 \\
\textbf{BERT4Rec} \cite{bert4rec}~\textsuperscript{(CIKM'19)}& 0.3098 & 0.1764 & 0.1357 & 0.0760 & 0.0393 & 0.0282 & 0.1053 & 0.0538 & 0.0381 \\
\midrule
\textbf{Mamba4Rec} \cite{Mamba4Rec}~\textsuperscript{(KDD'24)} & 0.3121 & \cellcolor{blue}\underline{\textit{0.1822}} & \cellcolor{blue}\underline{\textit{0.1376}} & 0.0812 & \cellcolor{blue}\underline{\textit{0.0461}} & \cellcolor{red}\textbf{0.0362} & 0.1152 & \cellcolor{blue}\underline{\textit{0.0603}} & \cellcolor{blue}\underline{\textit{0.0438}} \\
\textbf{SIGMA} \cite{SIGMA} ~\textsuperscript{(AAAI'25)}& \cellcolor{blue}\underline{\textit{0.3136}}  & 0.1777 & 0.1361& 0.0678 &0.0376&0.0284 &\cellcolor{blue}\underline{\textit{0.1239}} &0.0592 &0.0430\\
\midrule
\textbf{DSRec$^\star$} &\cellcolor{red}\textbf{0.3217$^\star$}&\cellcolor{red}\textbf{0.1869$^\star$}&\cellcolor{red}\textbf{0.1454$^\star$}&\cellcolor{red}\textbf{0.0871$^\star$}&\cellcolor{red}\textbf{0.0508$^\star$}&\cellcolor{blue}\underline{\textit{0.0354}}&\cellcolor{red}\textbf{0.1287$^\star$}&\cellcolor{red}\textbf{0.0661$^\star$}&\cellcolor{red}\textbf{0.0452$^\star$}\\
\#Improve  &2.58\% &2.57\%&5.67\%&2.83\%&10.19\%&-2.2\%&3.87\%&9.61\%&3.20\%\\
\bottomrule
\end{tabular}
\label{tab: overall}
\end{table*}

\vspace{2mm}

Table~\ref{tab: overall} summarizes the overall performance of our proposed \textbf{DSRec} model compared to a wide range of state-of-the-art sequential recommendation baselines. Best results are highlighted in \textbf{bold red}, while the second-best results are \underline{\textit{underlined in blue}}. From the table, we can make the following  observations:

\begin{itemize}
\item Our DSRec achieves state-of-the-art performance across all datasets and most evaluation metrics. On \textbf{MovieLens-1M}, DSRec outperforms the best-performing baseline (SIGMA \cite{SIGMA}) by $+2.58\%$ in HR@10, while achieving a notable $+5.67\%$ gain in MRR@10 over Mamba4Rec. These improvements demonstrate the advantage of our dual-interest embedding design and residual cross-SSM in capturing semantic signals in dense user sequences.
On the more sparse datasets \textbf{Amazon-Beauty} and \textbf{Amazon-Video-Games}, DSRec continues to outperform other models, validating its effectiveness in data sparsity scenario.
\item Our experimental results indicate that Transformer-based models outperform RNNs and CNNs based methods, in sequential recommendation tasks. This enhanced performance can be attributed to the Transformer's ability to process sequences in parallel and capture complex dependencies over long distances, which is particularly advantageous for modeling user-item interactions.
\item SSM-based models (Mamba4Rec \cite{Mamba4Rec}, SIGMA \cite{SIGMA}) demonstrate further improvements over Transformers, especially on datasets like MovieLens-1M with longer sequence lengths. This aligns with recent findings \cite{Mamba, Jamba} that SSMs can efficiently capture long-range dependencies while maintaining linear scalability.
The performance gap between SSM and Transformer models is more significant in longer sequence than in short sessions, validating the architectural design of structured temporal modeling.
\end{itemize}

\vspace{2mm}

\subsection{Ablation Study}

\begin{table*}[ht]
\centering
\setlength{\tabcolsep}{3mm}
\caption{Ablation study on Model Variant. \textbf{Bold} data indicate the best results.}
\label{tab:ablation}
\begin{tabular}{lcc|cc|cc}
\toprule
\multirow{2}{*}\textbf{Model Variant} & \multicolumn{2}{c|}{\textbf{MovieLens-1M}} & \multicolumn{2}{c|}{\textbf{Amazon-Beauty}} & \multicolumn{2}{c}{\textbf{Amazon-Video-Games}} \\
~& \textbf{HR@10} & \textbf{NDCG@10}  & \textbf{HR@10} & \textbf{NDCG@10} & \textbf{HR@10} & \textbf{NDCG@10} \\
\midrule
\textbf{DSRec} (full)       &\colorbox{red}{\textbf{0.3217}} &\colorbox{red}{\textbf{0.1869}}&\colorbox{red}{\textbf{0.0871}}&\colorbox{red}{\textbf{0.0508}}&\colorbox{red}{\textbf{0.1287}}&\colorbox{red}{\textbf{0.0661}}\\
\midrule
w/o Residual Cross-Fusion    &0.3180 & 0.1755 & 0.0867 &0.0433 & 0.1271 &  0.0638 \\
w/o Dual interest embedding (single)  &0.2983  &0.1710 &  0.0726 &0.0448 &  0.1172 &  0.0656 \\
w/o SSM\_S  & 0.2346 &0.1329  &0.0601  &0.0360 &0.0911  &0.0499  \\
w/o Fusion Encoder    &0.3134  &0.1868   &0.0655 &0.0348  &0.1164 &0.0617 \\
\midrule
Dual Mamba &0.3200  &0.1855  &0.0665  &0.0382 &0.1093  &0.0592  \\
\bottomrule
\end{tabular}
\label{tab: ablation}
\end{table*}

\vspace{2mm}

To validate the contributions of each component in DSRec, we conduct comprehensive ablation studies by systematically removing (w/o) or modifying key modules. The results are presented in Table~\ref{tab: ablation}, evaluated on three datasets and reported by HR@10 and NDCG@10.

\vspace{2mm}

\paragraph{Impact of Residual Cross-Fusion}  
From the comparison results, we can observe that removing the residual cross-fusion module significantly degrades performance on all datasets, especially on MovieLens-1M and Amazon-Beauty. This demonstrates the importance of inter-branch communication, where each interest embedding branch (long-term and short-term) benefits from contextual signals in the other.

\vspace{2mm}

\paragraph{Impact of Dual interest embedding Representation}  
Replacing the dual-interest embedding input with a single-interest embedding, as other baselines do, results in notable performance degradation across all benchmarks. This confirms our hypothesis that items assume different semantic roles in short-term and long-term contexts, and a unified interest embedding is insufficient to capture such polysemy behavior.

\vspace{2mm}

\paragraph{Impact of SSM-S Branch}  
Removing the short-term SSM-S encoder while retaining the long-term SSM-L leads to a notable performance drop, especially on the MovieLens dataset. This suggests that the short-term intent modeling is vital, particularly in sequential data where recent actions carry strong semantic signals.

\vspace{2mm}

\paragraph{Impact of Fusion Encoder}  
Without the fusion encoder, the model retains the dual-interest embedding structure but lacks effective integration before output. While the impact is slight, the drop in NDCG@10 on Amazon datasets indicates its importance in combining interest-specific outputs for fine-grained ranking.

\vspace{2mm}

\paragraph{Dual Mamba Baseline}  
We also construct a variant with two identical Mamba modules (Dual Mamba) without semantic-specific design. While this variant performs competitively on MovieLens-1M, it underperforms significantly on Amazon-Beauty and Amazon-Video-Games. This observation confirms that our asymmetric dual-SSM design is more effective than merely duplicating Mamba blocks.

\vspace{2mm}

These ablation observations validate the necessity of each proposed component. The dual-interest embedding, heterogeneous SSM encoders, and residual cross-fusion are all crucial for capturing multi-granular temporal dynamics and achieving state-of-the-art performance.

\subsection{Hyperparameter Analysis}
\begin{figure}[t]
  \centering
  \setlength{\abovecaptionskip}{0.2cm}
  \includegraphics[width=\linewidth]{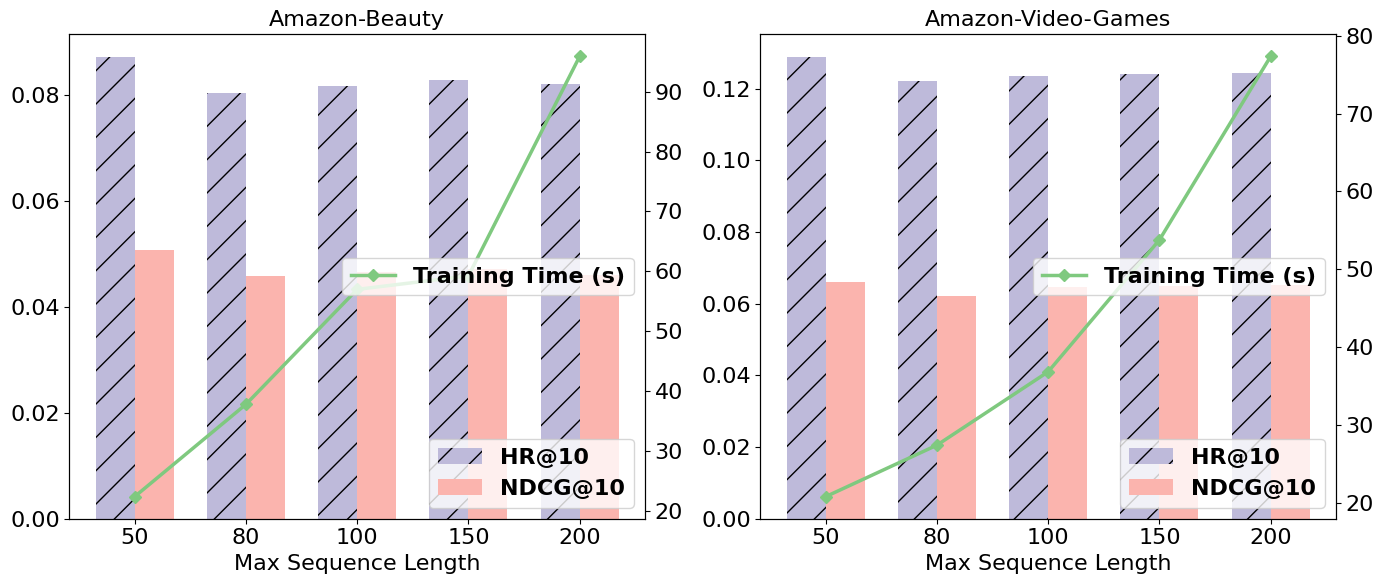}
  \caption{Max Sequence Length $T$ study on Amazon-Beauty and Amazon-Video-Games.}
  \label{fig: hyper_L}
\end{figure}

In this section, we conduct experiments on Beauty to analyze the influence of two significant hyperparameters: \textit{a)} L, the maximum sequential length; \textit{b)} B, the number of stacked DSRec blocks. The results are respectively visualized in Fig.~\ref{fig: hyper_L} and Table \ref{tab: hyper_B}.

\vspace{2mm}

\vspace{2mm}

\begin{table}[h]
\centering
\setlength{\tabcolsep}{1.5mm}
\caption{Parameter study for $B$ on Amazon-Beauty and Amazon-Video-Games.}
\begin{tabular}{lccccc}
\toprule
\textbf{Dataset} &\textbf{\#blocks} & \textbf{HR@10} & \textbf{NDCG@10} & \textbf{Training(s)} & \textbf{GPU Mem} \\ 
\midrule
\multirow{5}{*}{\textbf{Beauty}}
&1 & 0.0849  &0.0524  & 12.90s & 2.87 G \\ 
&2 &0.0861  &0.0507 & 17.71s & 3.55 G \\ 
&3 & 0.0871 & 0.0508 &22.41s  &4.24 G  \\
&4 &0.0786  &0.0422 & 23.37s & 5.62 G \\ 
&5 &0.0777  &0.0411 & 25.76s &6.59 G  \\ 
\midrule
\multirow{5}{*}{\textbf{Games}}
&1 &0.1227 &0.0634 &15.05s &2.60 G \\
&2 &0.1287 &0.0661 &20.83s &3.47 G \\
&3 &0.1221 &0.063 &21.50s &4.44 G \\
&4 &0.1194 &0.0599 &29.07s &5.41 G \\
&5 &0.1178 &0.0600 &30.21s &6.39 G \\
\bottomrule
\end{tabular}
\label{tab: hyper_B}
\end{table}

\vspace{2mm}

\paragraph{Effect of the Number of Blocks $B$}  
From Table~\ref{tab: hyper_B}, we observe that on Amazon-Beauty and Amazon-Video-Games, the model achieves the highest HR@10 and NDCG@10 when $B=3$ and $B=2$, respectively. This suggests that stacking multiple residual-coupled dual-SSM layers improves to capture multi-granular temporal dynamics. However, further increasing $B$ beyond 3 leads to degradation in performance. This is likely due to overfitting or optimization difficulties caused by deeper architectures, especially on sparse datasets. Moreover, we observe a steady increase in training time and GPU memory consumption with more layers, highlighting a trade-off between performance and efficiency.

\vspace{2mm}

\paragraph{Effect of Maximum Sequence Length $T$}  
Fig. ~\ref{fig: hyper_L} shows the performance varying the maximum sequence length. On both Amazon-Beauty and Amazon-Video-Games, we observe a peak at $T=50$, after which performance gradually drops. This suggests that for sparse and short-session datasets, incorporating overly long histories may introduce noise or dilute recent preference signals. Furthermore, longer sequences also lead to substantial computational overhead, as training time nearly quadruples from $T=50$ to $T=200$.
These results highlight the importance of tuning both structural depth and temporal context window to match dataset characteristics. DSRec demonstrates strong adaptability across a wide range of hyperparameter settings.


\vspace{2mm}

\subsection{Model Complexity and Efficiency}

We analyze the theoretical and empirical complexity of DSRec compared to Transformer- and SSM-based baselines. In addition, we also compare prediction and training efficiency with the most competitive baseline.

\paragraph{Theoretical Complexity}
As shown in Table~\ref{tab:complexity}, the Transformer-based models incur quadratic time and space cost with respect to sequence length $T$, making them less suitable for long sequences. RNNs and SSMs maintain linear scaling in $T$. While both RNNs and SSMs require $\mathcal{O}(d^2)$ operations due to dense linear transitions, SSMs can still outperform RNNs in parallelizability and hardware efficiency. DSRec maintains linear complexity in sequence length but doubles the cost in hidden size due to its dual-branch architecture, which remains attractive and scalable in practice.
\begin{table}[ht]
\centering
\caption{Theoretical time complexity comparison. $T$: sequence length, $d$: dimension of hidden layers.}
\label{tab:complexity}
\begin{tabular}{lcc}
\toprule
\textbf{Model} & \textbf{Time Comp.} \\
\midrule
RNN (GRU4Rec, NARM) & $\mathcal{O}(T \cdot d^2)$ \\
Transformer (SASRec, BERT4Rec) & $\mathcal{O}(T^2 \cdot d)$ \\
SSM (Mamba4Rec, SIGMA, \textbf{DSRec}) & $\mathcal{O}(T \cdot d^2)$ \\
\bottomrule
\end{tabular}
\label{tab:complexity}
\end{table}

\begin{figure}[t!]
  \centering
  \setlength{\abovecaptionskip}{0.2cm}
  \includegraphics[width=\linewidth]{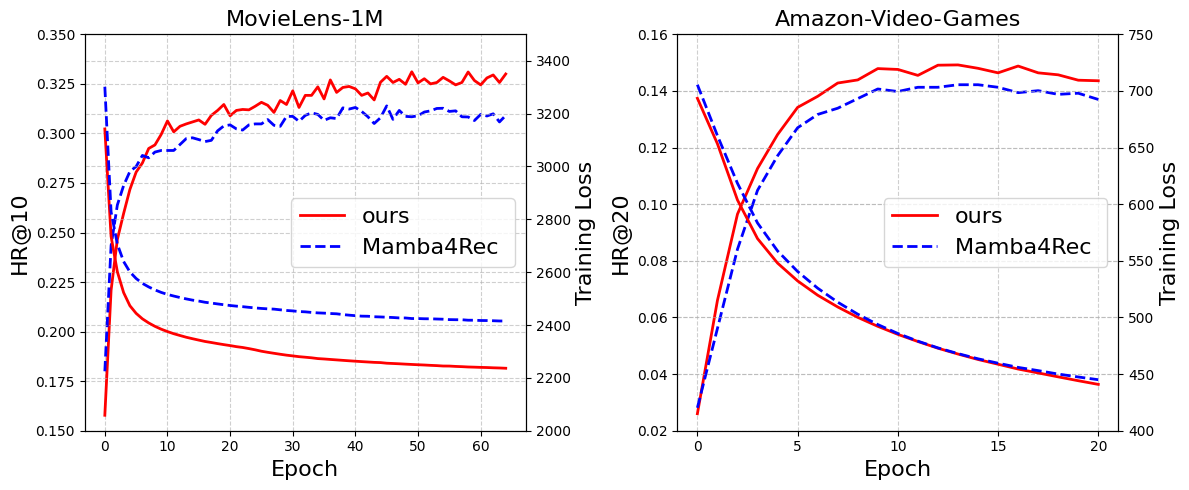}
  \caption{Prediction comparison and training loss comparison on MovieLens-1M and Amazon-Video-Games datasets.}
  \label{fig: loss}
\end{figure}

\vspace{2mm}

\paragraph{Prediction and Training Efficiency}
Fig. \ref{fig: loss} presents a comparative analysis of model performance, focusing on HR@10 and training loss across epochs for two models: Mamba4Rec and DSRec, evaluated on the MovieLens-1M and Amazon-Video-Games datasets.
The comparative analysis shows that DSRec generally outperforms Mamba4Rec in terms of HR@10, and faster initial convergence in training loss. This suggests that DSRec is better suited for capturing complex patterns in sequential recommendation tasks, particularly on long sequential dataset.

\vspace{2mm}

\paragraph{Runtime Efficiency}
We further compare wall-clock inference time and GPU memory cost at Table \ref{tab: cost}. The experimental results highlight SSM's superior runtime efficiency, with both faster inference times and lower GPU memory.
Although DSRec introduces two SSM encoders, both are linear-time and lightweight. Thus, the overall inference costs remain close to SSM baselines and are significantly more efficient than Transformer-based method.
\begin{table}[ht]
\centering
\caption{Inference time and GPU Memory cost comparisoneachepoch (Amazon-Beauty, batch=2048).}
\label{tab:runtime}
\begin{tabular}{lcc}
\toprule
\textbf{Model}  & \textbf{Inference Time} & \textbf{GPU Mem.}\\
\midrule
SASRec      &123ms  &7.58G  \\
\midrule
Mamba4Rec      &72ms  & 2.69G  \\
\textbf{DSRec} &83ms & 3.55G \\
\bottomrule
\end{tabular}
\label{tab: cost}
\end{table}

\vspace{2mm}

\subsection{Robustness Analysis}
Recommender systems often suffer from the cold-start and long-tail problems, where items with few interactions are poorly represented during training.
To assess the robustness of DSRec across users with varying historical interaction lengths, we partition the test users into three groups based on the number of their training interactions: 
\textit{0–5}, \textit{5–20}, and \textit{20+}. Each subset corresponds to users with sparse, moderate, and rich historical behavior, respectively. We compare our model DSRec against the strong SSM-based baseline and report metrics including Hit@10/20, NDCG@10/20, and MRR@10/20 for each group.
\vspace{2mm}

\begin{figure}[ht]
  \centering
  \setlength{\abovecaptionskip}{0.2cm}
  \includegraphics[width=\linewidth]{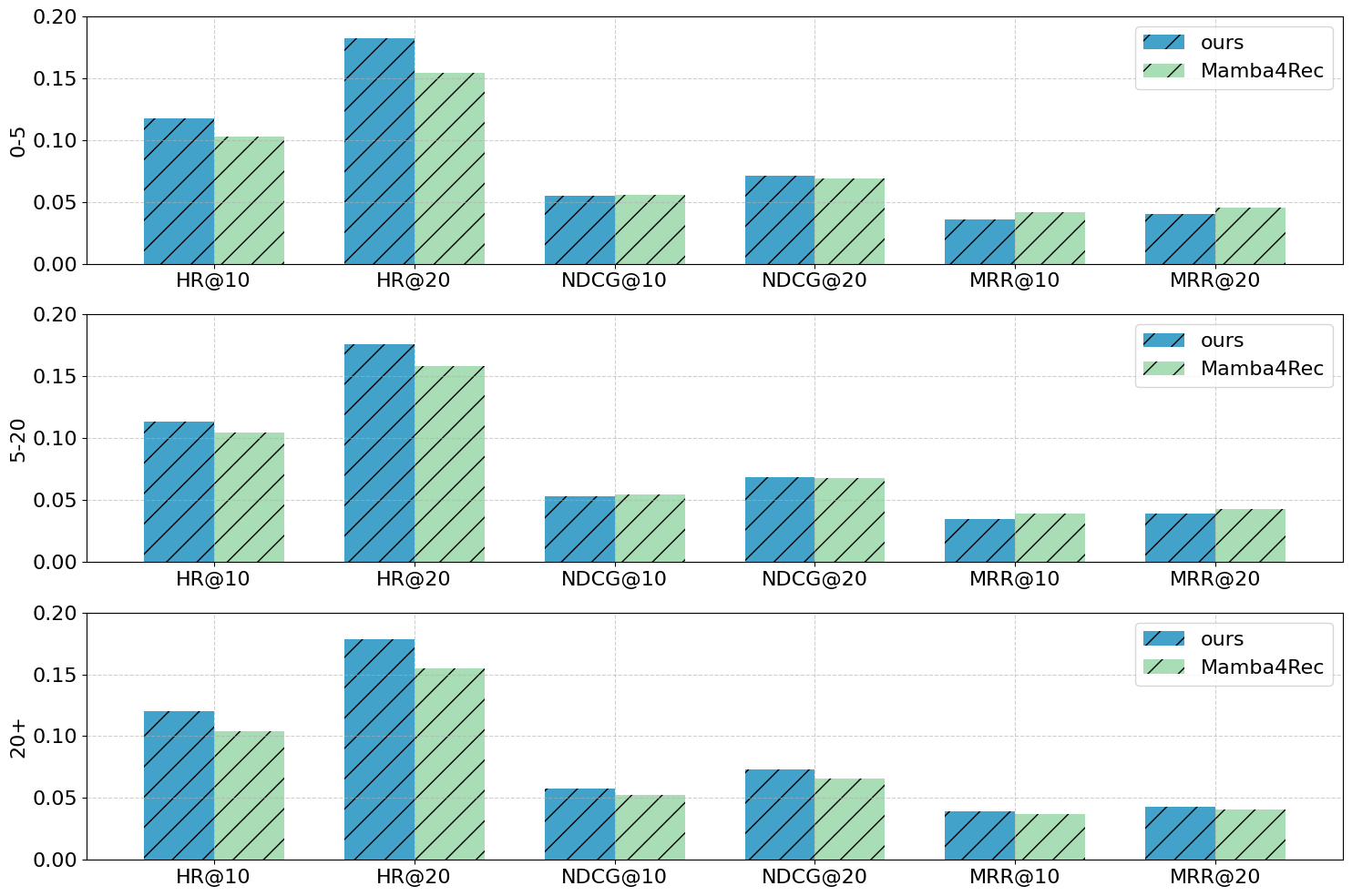}
  \caption{Robustness comparison across user history length groups in Amazon-Video-Games.}
  \label{fig: cold}
\end{figure}


From Fig. \ref{fig: cold}, we can observe that for \textit{cold-start} users with short histories (0–5 interactions), DSRec significantly outperforms Mamba4Rec, which indicating superior modeling of early-stage preferences.
For mid-range users (5–20), DSRec continues to outperform in most metrics, suggesting its ability to balance long- and short-term signals.
For users with rich histories (20+), DSRec achieves the highest performance, confirming the benefit of its dual-interest encoding for long-horizon modeling. These results demonstrate the robustness of DSRec under varying interaction sparsity. The consistent improvements across all user groups validate the effectiveness of our DSRec.


\section{Conclusion and Future Work}
Most existing sequential recommendation models implicitly assuming that each item has consistent semantics across all user contexts. This limits their ability to capture the polysemous nature of item behavior. Moreover, standard state space architectures failing to account for the distinct dynamics of long-term preferences versus short-term session-based intent.
To address these challenges, in this paper, we propose \textbf{DSRec}, a dual-interest embedding state-space model for sequential recommendation that explicitly models both long-term and short-term user preferences. Our method introduces a semantic disentanglement mechanism where each item is projected into a long-term interest embedding and a short-term interest embedding. These embeddings are encoded separately by two specialized SSM encoders: a full-sequence Mamba and a time-modulated sensitive SSM to inter-click temporal gaps. To enable cross-granular coordination without semantic interference, we design a residual cross-fusion mechanism that injects context between branches.
Extensive experiments on multiple benchmark datasets demonstrate the effectiveness of DSRec. 

\vspace{2mm}
\noindent\textbf{Future Work.}  
While DSRec achieves advanced performance, several limitations remain open for future research. First, the item polysemous modeling is novel but basic, we plan to extend by context aware dual-tokenization to incorporate item-side semantic roles, such as Variational AutoEncoder (VAE). Second, sparse context information in transactions need the integration of external knowledge (such as graph structure) into the interest embedding, which may further enhance generalization. Finally, exploring a learnable fusion gate or contrastive alignment between the two semantic branches could yield more adaptive interactions beyond residual fusion. We also intend to test DSRec in cross-domain settings to assess its broader applications.

\section*{Acknowledgments}

\bibliographystyle{IEEEtran}
\bibliography{base}

\end{document}